\documentclass[conference]{IEEEtran}
\usepackage{times}

\usepackage[numbers]{natbib}
\usepackage{multicol}
\usepackage[bookmarks=true]{hyperref}
\usepackage{eucal}
\usepackage{amsmath}
\usepackage{amsthm}
\usepackage{amsfonts}
\usepackage{mathtools}
\usepackage{todonotes}
\usepackage{algorithm}
\usepackage{algpseudocode}
\usepackage{multicol}
\usepackage{subfigure}
\usepackage{dblfloatfix}

\usepackage[capitalise]{cleveref}
\crefformat{equation}{(#2#1#3)}
\crefrangeformat{equation}{(#3#1#4) to~(#5#2#6)}
\crefname{figure}{Figure}{Figures}

\newcommand{\algmargin}{\the\ALG@thistlm}
\makeatother
\newlength{\whilewidth}
\algdef{SE}[parWHILE]{parWhile}{EndparWhile}[1]
  {\parbox[t]{\dimexpr\linewidth-\algmargin}{%
     \hangindent\whilewidth\strut\algorithmicwhile\ #1\ \algorithmicdo\strut}}{\algorithmicend\ \algorithmicwhile}%
\algnewcommand{\parState}[1]{\State%
  \parbox[t]{\dimexpr\linewidth-\algmargin}{\strut #1\strut}}

\newtheorem{theorem}{Theorem}[section]

\input{irom.sty}

\IEEEoverridecommandlockouts

\begin{document}

\title{Learning Task-Driven Control Policies \\ via Information Bottlenecks}

\author{Vincent Pacelli and Anirudha Majumdar
\thanks{The authors are with the Mechanical and Aerospace Engineering department at Princeton University, NJ, 08540, USA
        {\tt\small \{vpacelli, ani.majumdar\}@princeton.edu}%
}%
}

\maketitle

\begin{abstract}
This paper presents a reinforcement learning approach to synthesizing \emph{task-driven} control policies for robotic systems equipped with rich sensory modalities (e.g., vision or depth). Standard reinforcement learning algorithms typically produce policies that tightly couple control actions to the \emph{entirety} of the system's state and rich sensor observations. As a consequence, the resulting policies can often be sensitive to changes in \emph{task-irrelevant} portions of the state or observations (e.g., changing background colors). In contrast, the approach we present here learns to create a task-driven representation that is used to compute control actions.  Formally, this is achieved by deriving a policy gradient-style algorithm that creates an \emph{information bottleneck} between the states and the task-driven representation; this constrains actions to only depend on task-relevant information. We demonstrate our approach in a thorough set of simulation results on multiple examples including a grasping task that utilizes depth images and a ball-catching task that utilizes RGB images. Comparisons with a standard policy gradient approach demonstrate that the task-driven policies produced by our algorithm are often significantly more robust to sensor noise and task-irrelevant changes in the environment. 

\end{abstract}

\IEEEpeerreviewmaketitle

\section{Introduction}
\label{sec:intro}
The increasing availability of high-resolution sensors has significantly contributed to the recent explosion of robotics applications. For example, high-precision cameras and LIDARs now allow autonomous vehicles to perform tasks ranging from navigating busy city streets to mapping mines and buildings. By far the most common approach adopted by such systems is to utilize as much sensor information as is available in order to estimate the full state of these complex environments; the resulting state estimates are then used by the robot to choose control actions necessary to complete its task. 
However, this ubiquitous approach does not distinguish between the \emph{task-relevant} and \emph{task-irrelevant} portions of the system's state. The result is an unnecessarily tight coupling between sensor observations and control actions that is often not robust to noise or uncertainty in irrelevant portions of the state. Moreover, these policies tend to have a large computational burden. This computational requirement can manifest either as a state estimator that needs to process large amounts of data in real time, or as a policy learned from sampling a large number of diverse operating environments. The goal of this paper is to address these challenges by synthesizing control policies that only depend on \emph{task-driven representations} of the robot's state. Doing so reduces the coupling between the states and control actions, improves robustness, and reduces computational requirements. 

\begin{figure}[t]
\begin{center}
\includegraphics[width=0.99\columnwidth]{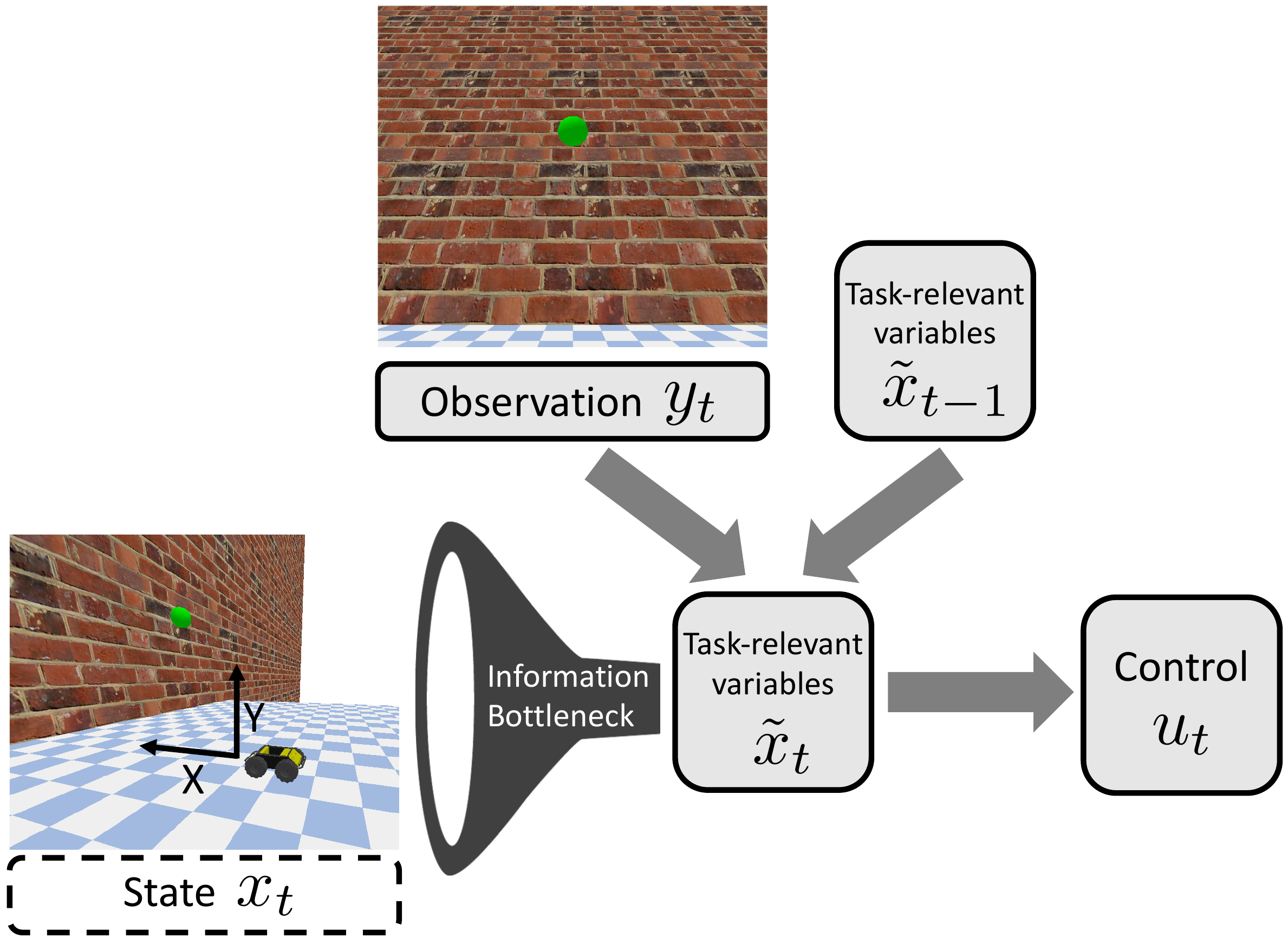}
\end{center} 
\vspace{-14pt}
\caption{A depiction of our approach applied to the ball-catching problem described in \cref{ssec:ball}. Our policy implements an information bottleneck that limits the amount of state information (in this case, the robot and ball position and velocities) extracted from the sensor observations (RGB images) to create a set of task-relevant variables (TRVs) on which the control action (robot velocity) depends. The result is a policy that is more robust to image noise and changes to the brick background texture.\label{fig:anchor}}
\vspace{-18pt}
\end{figure}

To illustrate the advantages of a task-driven policy, consider a ball-catching example. An agent equipped with a high-resolution camera is tasked with catching a ball (Figure \ref{fig:anchor}). Using its sensor, the agent can attempt to estimate the full state of the system (e.g., ball velocity, ego-motion, wind speed, etc.), feed this information into a physical model for the ball's flight, integrate the model to find where the ball will land, and move to this location to catch it. This approach requires estimating every parameter involved in the ball's motion and can easily be compromised when the ball is far away and difficult to see or when there are visual artifacts such as glare.  

Instead, extensive cognitive psychology experiments \cite{Gigerenzer07, McLeod03, Shaffer04} have demonstrated that humans employ a task-driven strategy known as the gaze heuristic to catch projectiles. This strategy simply modulates the human's speed in order to fix the position of the ball in their visual field. This policy, which only requires minimal sensor information and internal computation, naturally drives the human to the ball's landing position. The gaze heuristic highlights that task-driven control policies are often robust control policies that only depend on small amounts of salient state information. Specifically, the heuristic is naturally \emph{robust to distributional shifts} in irrelevant portions of the environment (e.g., the visual backdrop) while also being adaptable to perturbations like a gust of wind altering the ball's course. Traditionally, task-driven policies have required hand-engineering for each control problem. This process can be difficult and time consuming. The goal of this paper is to propose a reinforcement learning framework that \emph{automatically synthesizes} task-driven control policies for systems with nonlinear dynamics and high-dimensional observations (e.g., RGB or depth images). 


\textbf{Statement of Contributions.} The main technical contribution of this paper is to formulate a reinforcement learning algorithm that synthesizes  task-driven control policies. This synthesis is achieved by creating an \emph{information bottleneck} \cite{Tishby99} that limits how much state information the policy is allowed to use. We present a reinforcement learning algorithm --- referred to as task-driven policy gradient (TDPG) --- that leverages the recently-proposed mutual information neural estimator \cite{Belghazi18} to tractably search for an effective task-driven policy. Finally, we demonstrate that this formulation provides the key advantage of a task-driven approach --- robustness to perturbations in task-irrelevant state and sensor variables. This benefit is demonstrated in three examples featuring nonlinear dynamics and high-dimensional sensor models: an adaption of the lava problem from the literature on partially observable Markov decision processes (POMDPs), ball catching using RGB images, and grasping an object using depth images.  

\subsection{Related Work} 

{\bf State Estimation and Differentiable Filtering.} Classical approaches to controlling robotic systems typically involve two distinct pipelines: one that uses the robot's sensors to estimate its state and another that uses this state estimate to choose control actions. Such an architecture is motivated by the \emph{separation principle} \cite{Anderson07} from (linear) control theory and allows one to leverage powerful control-theoretic techniques for robust estimation and control \cite{Dullerud13, Zhou98}. A recent line of work on \emph{differentiable filtering} \cite{Haarnoja16, Jonschkowski18, Karkus18} has extended this traditional pipeline to elegantly handle rich sensor observations (e.g., images) by \emph{learning} state estimators in an end-to-end manner via deep learning. However, as the gaze heuristic example from Section \ref{sec:intro} demonstrates, full state representations are often overly rich when viewed from the perspective of the task at hand. This is particularly true in settings where representing the full state requires capturing the state of the robot's environment. Instead, our goal is to learn minimalistic \emph{task-driven} representations that are sufficient for control. Such representations can be highly compact as compared to full state representations. Moreover, carefully-constructed task-driven representations have the potential to be robust to sensor noise and changes to irrelevant portions of the robot's environment. Intuitively, this is because uncertainty or noise in irrelevant portions of the sensor observations are filtered out and thus no longer corrupt the robot's actions. We present a theoretical result in \cref{sec:recurrent} along with simulation experiments in \cref{sec:examples} to support this intuition. 

%
%
%

{\bf End-to-end Learning of Policies.} Deep reinforcement learning approaches have the ability to learn control policies in an \emph{end-to-end} manner \cite{Gupta17, Gupta17a, Karkus19, Levine16, Levine18, Sunderhauf18, Zhu17}. Such end-to-end approaches learn to create representations that are tuned to the task at hand by exploiting statistical regularities in the robot's observations, dynamics, and environment. However, these methods do not explicitly attempt to learn representations that are task-driven (i.e., representations that filter out portions of the sensor observations that are irrelevant to the task). As a result, policies trained via standard deep RL techniques may be sensitive to changes in irrelevant portions of the robot's environment (e.g., changes to the background color in a ball-catching task). In contrast, the deep RL-based approach we present seeks to explicitly learn task-driven representations that filter out irrelevant factors. Our simulation results in Section \ref{sec:examples} empirically demonstrate that our approach is robust to such \emph{distributional shifts}.

{\bf Information Bottlenecks.} Originally developed in the information theory literature, \emph{information bottlenecks} \cite{Tishby99, Alemi16, Kolchinsky19, Tishby15} allow one to formalize the notion of a ``minimal-information representation" that is sufficient for a given task (e.g., a prediction task in the context of supervised learning). Given an input random variable $X$ and a target random variable $Y$, one seeks a representation $\tilde{X}$ that forms a Markovian structure $X \rightarrow \tilde{X} \rightarrow Y$. The representation is chosen to minimize the mutual information between $X$ and $\tilde{X}$ while still maintaining enough information to predict $Y$ from $\tilde{X}$. 
Recent work has sought to adapt this theory for synthesizing task-driven representations for control \cite{Pacelli19, Achille18}. In \cite{Pacelli19},  a model-based approach for automatically synthesizing task-driven representations via information bottlenecks is presented. However, this approach is limited to settings with an explicit model of the robot's sensor and dynamics. This prevents the approach from being applied to systems with rich sensing modalities (e.g. RGB or depth images), for which one cannot assume a model. In contrast, the approach we present here \emph{learns} to create a task-driven representation via reinforcement learning and is directly applicable to settings with rich sensing. In \cite{Achille18}, the authors use information bottlenecks to define minimal state representations for control tasks involving high-dimensional sensor observations (this approach is also related to the notion of \emph{actionable information} \cite{Soatto11, Soatto13} in vision; see \cite{Achille18} for a discussion). However, \cite{Achille18} does not present concrete algorithms for learning task-driven representations. In contrast, we present  a policy gradient algorithm based on \emph{mutual information neural estimation} (MINE) \cite{Belghazi18}. Moreover, we note that our definition of a task-driven policy differs from that of \cite{Achille18}; the representations we learn seek to create a bottleneck between sensor observations and control actions as opposed to finding a minimal representation for predicting costs. 


{\bf Exploration in RL via Mutual Information Regularization.} The information bottleneck principle has also been used to improve sample efficiency in RL by encouraging exploration \cite{Goyal18, Yingjun19}. More generally, there is a line of work on endowing RL agents with \emph{intrinsic motivation} by maximizing the mutual information between the agent's control actions and states \cite{Klyubin05, Salge13, Salge14, Tiomkin17, Yu19, Zhao19}. Our focus instead is on improving robustness and generalization of policies; the information bottleneck-based approach we present is aimed at learning task-driven representations for control. This is achieved by minimizing the mutual information between the agent's states and learned task-relevant variables.

\section{Defining Task-Driven Policies}
\label{sec:recurrent}
In this section, we formalize our notion of a task-driven control policy. Our definition is in terms of a reinforcement learning problem whose solution produces a set of task-relevant variables (TRVs) and a policy that depends only on these variables. We begin by formulating the problem as a finite-horizon partially-observable Markov decision process (POMDP) \cite{Sutton98}. The robot's states, control actions, and sensor observations at time $t \in \{0, \dots, T\}$ are denoted by $x_t \in \mathcal{X}_t,\ u_t \in \mathcal{U}_t$, and $y_t \in \mathcal{Y}_t$ respectively. The robot dynamics and sensor model are denoted by the conditional distributions $p(x_{t + 1} | x_t, u_t)$ and $s(y_t | x_t)$ respectively. We do not assume knowledge of either of these distributions. The functions $c_0(x_0, u_0), \dots, c_{T - 1}(x_{T - 1}, u_{T - 1})$ describe the cost at each time step with a terminal cost specified by $c_T(x_t, u_t) \coloneqq c_T(x_T)$. Our goal is to find a policy $\pi_t(u_t | y_t)$ that solves
\begin{align}
	\underset{\pi_t(u_t|y_t)}{\text{minimize}}\ \EE\left[c(\tau)\right] \coloneqq \EE\left[\sum_{t = 0}^{T}c_t\right], \label{eq:pomdp}
\end{align}
when run online in a test environment. Throughout this paper, we use $\EE[\cdot]$ to denote the expectation; it is subscripted with a distribution when necessary for clarity. 

To achieve our goal of learning a task-driven policy, we choose the policy to have the recurrent structure illustrated in Figure \ref{fig:anchor}. We refer to $\tilde{x}_t \in \tilde{\mathcal{X}}_t$ as the \emph{task-relevant variables} (TRVs) and search for two conditional distributions: $q(\tilde{x}_t | \tilde{x}_{t - 1}, y_t), \pi(u_t | \tilde{x}_t)$. The first specifies a (stochastic) mapping from the previous TRVs and the current observation to the current TRVs. The second conditional distribution computes the control actions given the current TRVs. These will be parameterized by neural networks in the following sections. The overall policy can thus be expressed as:
\begin{align}
	\pi(u_t | y_t) = \int_{\tilde{\mathcal{X}}_t \times \tilde{\mathcal{X}}_{t - 1}} \hspace{-5mm} \pi(u_t | \tilde{x}_t) q(\tilde{x}_t | \tilde{x}_{t - 1}, y_t) q_t(\tilde{x}_{t-1})\ \mathrm{d}(\tilde{x}_t, \tilde{x}_{t - 1}). \nonumber 
\end{align}
Our goal is to learn TRVs that form a \emph{compressed representation} of the state that is sufficient for the purpose of control. To formalize this, we leverage the theory of information bottlenecks \cite{Tishby99} to limit how much information $\tilde{x}_t$ contains about $x_t$. This is quantified using the mutual information
\begin{align}
\mathbb{I}[x_t; \tilde{x}_t] \coloneqq \mathbb{D}\left[p_t(x_t, \tilde{x}_t)\| p_t(x_t)q_t(\tilde{x}_t)\right] \label{eq:mi}
\end{align}
between the state $x_t$ and TRVs $\tilde{x}_t$ at each time step. Here, $\mathbb{D}[\cdot\|\cdot]$ is the Kullback-Leibler (KL) divergence \cite{Cover12}. Intuitively, minimizing the mutual information corresponds to learning TRVs that are as independent of the state as possible. Thus, the TRVs ``filter out" irrelevant information from the state. Therefore, we would like to find a policy that solves:
\begin{align}
	\underset{\substack{q(\tilde{x}_t | \tilde{x}_{t - 1}, y_t)\\ \pi_t(u_t|\tilde{x}_t)}}{\text{minimize}}\ \mathcal{J} \coloneqq \beta \EE\left[c(\tau)\right] + \sum_{t = 0}^T\mathbb{I}[x_t; \tilde{x}_t]. \label{eq:robust pomdp}
\end{align}
Here, $\beta$ can be viewed as a Lagrange multiplier, and the above problem can be interpreted as minimizing the total information contained in $\tilde{x}_t$ subject to an upper bound on the expected cost over the time horizon.  In Section \ref{sec:algorithm}, we develop a reinforcement learning algorithm for tackling \cref{eq:robust pomdp}. We refer to the resulting policies as \emph{task-driven policies}. Note that the mutual information is invariant under bijective transforms of the random variables for which it is computed \cite{Belghazi18, Cover12}. As a result, the optimum value of $\mathcal{J}$ remains unchanged for equivalent representations of the robot and its environment.

Similar to the gaze heuristic (ref. Section \ref{sec:intro}), we expect that a task-driven policy as defined above will be a robust policy. This benefit is conferred to our policy by minimizing the mutual information between states and TRVs. Intuitively, a policy is closer to being open-loop when less state information is present in $\tilde{x}_t$. The more open-loop the policy is, the less it is impacted by changes in the state or sensor distributions between environments. In \cite{Pacelli19}, this intuition is formalized using the theory of \emph{risk metrics} with the following theorem:
\begin{theorem}
\label{thm:robustness}
Define the \emph{entropic risk metric} \cite[Example 6.20]{Shapiro09}:
\begin{equation}
\label{eq:entropic risk}
\rho_\beta \left[ c_t \right] \coloneqq \frac{1}{\beta} \log \Big{[} \EE_{p_t} \exp(\beta c_t)\Big{]}.
\end{equation}
Let $\check{p}_t(x_t, \tilde{x}_t, u_t)$ be any distribution satisfying:
\begin{align}
\label{eq:KL condition}
&\beta \KL[\check{p}_t(x_t, \tilde{x}_t, u_t) \| p_t(x_t, \tilde{x}_t, u_t)]\\
&\leq \KL[p_t(x_t, \tilde{x}_t) \| p_t(x_t) q_t(\tilde{x}_t)]. \nonumber
\end{align}
Then, the online expected cost is bounded by a combination of the entropic risk and mutual information:
\begin{equation}
\label{eq:cost bound}
\EE_{\check{p}_t}\left[c(\tau)\right] \leq \sum_{t=0}^{T} \Big{(} \rho_\beta \left[ c_t \right] + \mathbb{I}[x_t; \tilde{x}_t] \Big{)}.
\end{equation}
\vspace{-5pt}
\end{theorem}
The entropic risk is a functional that is similar to the expectation, but also accounts for higher moments of the distribution, e.g. its variance. The parameter $\beta$ controls how much the metric weights the expected cost versus the higher moments, and $\lim_{\beta \to 0} \rho_\beta[c_t] = \EE[c_t]$. Optimizing $\rho_\beta$ can be difficult because computing its gradient requires computing $q(\tilde{x}_t)$ at each time step. Therefore, we optimize $\mathcal{J}$, a first-order approximation of \cref{eq:cost bound}. In section \cref{sec:examples}, we demonstrate in multiple examples that minimizing our objective $\mathcal{J}$ produces a robust policy that generalizes beyond its training environment.

\section{Learning Task-Driven Policies}
\label{sec:algorithm}
This section discusses finding a policy that approximately solves \eqref{eq:robust pomdp} within a reinforcement learning (RL) framework. 
If the mutual information term is removed from the objective $\mathcal{J}$, standard RL techniques such as policy gradient (PG) (e.g. \cite{Schulman15, Schulman17}) would be sufficient. However, the mutual information and its gradient are known to be difficult quantities to estimate. A number of tractable upper and lower bounds have been proposed recently to provide means for optimizing objectives containing a mutual information term \cite{Poole19}. We elected to use the recently-proposed mutual information neural estimator (MINE) \cite{Belghazi18} due to its accuracy and ease of implementation. We then derive a PG-style algorithm in \cref{ssec:pg}.

\subsection{Mutual Information Neural Estimator (MINE)}

The MINE is based on the Donsker-Varadhan (DV) variational representation of the KL divergence \cite[Theorem 2.3.2]{Gray11}. For any two distributions $P, Q$ defined on sample space $\Omega$, the KL divergence between them can be expressed as:
\begin{align}
	\KL[P\|Q] = \sup_{F: \Omega \to \RR}	 \mathcal{J}_{dv} \coloneqq \EE_P F - \log \EE_Q [\exp(F)]. \label{eq:dv}
\end{align}

The supremum is taken over all functions $F$ such that the two expectations are finite. Let $\hat{\EE}^N_P$ denote the empirical expectation computed with $N$ i.i.d. samples from $P$. We can estimate the KL divergence by replacing the function class over which the supremum is taken by a family of neural networks $F^\theta$ parameterized by $\theta \in \Theta$:
\begin{align}
	\hat{\KL}^N[P\|Q] \coloneqq \sup_{\theta \in \Theta}	 \underbrace{\hat{\EE}^N_P F^\theta - \log \hat{\EE}^N_Q \left[\exp\left(F^\theta\right)\right]}_{\hat{\mathcal{J}}_{dv}}.\label{eq:kl mine}
\end{align}
This approximation is a \emph{strongly consistent} underestimate of \cref{eq:dv} (see \cite{Belghazi18}). In short, this means that through choice of appropriate network structure and sample size, $\hat{\KL}^N[P\|Q]$ can approximate $\KL[P\|Q]$ arbitrarily closely. 

Since the mutual information is a KL divergence, we can approximate \cref{eq:mi} using the above KL approximation as:
\begin{align}
	\hat{\mathbb{I}}^N[x_t; \tilde{x}_t] \coloneqq \hat{\KL}^N\left[p_t(x_t, \tilde{x}_t)\| p_t(x_t)q_t(\tilde{x}_t)\right].
\end{align}
This is the MINE. The algorithm for computing the MINE estimate is presented in \cref{alg:mine}. At a high-level, the algorithm attempts to find neural network parameters $\theta$ in order to maximize $\hat{\mathcal{J}}_{dv}$ in \eqref{eq:kl mine} via stochastic gradient descent. 
We use the notation $\EE_\perp[\cdot]$ to denote the expectation taken using $p_t(x_t)q_t(\tilde{x}_t)$. The expectation computed using the joint distribution $p_t(x_t, \tilde{x}_t)$ is unsubscripted. These expectations are approximated by sampling two minibatches of size $B < N$ uniformly from the training batch. The minibatches are denoted in \cref{alg:mine} by indicies $j_1, \dots, j_B$ and $m_1, \dots, m_B$ respectively. The gradient of $\hat{\mathcal{J}}_{dv}$ is
\begin{align}
	\nabla_\theta \hat{\mathcal{J}}_{dv} = \hat{\mathbb{E}}^B\left[\nabla_\theta F^\theta\right]	 - \frac{\hat{\mathbb{E}}^B_\perp\left[\nabla_\theta F^\theta \exp\left(F^\theta\right)\right]}{\hat{\mathbb{E}}^B_\perp[\nabla_\theta \exp\left(F^\theta\right)]}. \label{eq:grad mine}
\end{align}
This gradient, which is computed using the minibatches, is used to update $\theta$ with stochastic gradient descent (or a similar optimizer like ADAM \cite{Kingma14}). For additional details on training the MINE, see \cite{Belghazi18}.

\begin{algorithm}[t]
\caption{Mutual Information Neural Estimator (MINE)}
\label{alg:mine}
\begin{algorithmic}[1]
	\Procedure{Train-Mine}{$\theta, \{(x^n, \tilde{x}^n)\}_{n = 1}^N$}
	\Repeat
		\State Sample joint minibatch: $\{(x^{j_b}, \tilde{x}^{j_b})\}_{b = 1}^B$.\
		\State Sample marginal minibatch: $\{\tilde{x}^{m_b}\}_{b = 1}^B$.\
		\State Compute $\hat{\mathbb{E}}^B\left[\nabla_\theta F^\theta\right]$ with $\{(x^{j_b}, \tilde{x}^{j_b})\}_{b = 1}^B$.\
		\State Compute $\frac{\hat{\mathbb{E}}^B_\perp\left[\nabla_\theta F^\theta\right]}{\hat{\mathbb{E}}^B_\perp[\nabla_\theta \exp\left(F^\theta\right)]}$ with $\{(x^{j_b}, \tilde{x}^{m_b})\}_{b = 1}^B$.
		\State Update $\theta$ using minibatches and \cref{eq:grad mine}.
	\Until{Convergence of $\hat{\mathcal{J}}_{dv}$.}\\
	\Return $\theta$
	\EndProcedure
	\end{algorithmic}
\end{algorithm}

\subsection{Task-Driven Policy Gradient Algorithm}
\label{ssec:pg}

We now describe our procedure for leveraging MINE within a policy gradient (PG) algorithm for tackling \eqref{eq:robust pomdp}. As described in Section \ref{sec:recurrent}, the policy is parameterized by two neural networks: a recurrent network $q^\phi(\tilde{x}_t | \tilde{x}_{t - 1}, y_t)$ that outputs the TRVs and a feedforward network  $\pi^\psi(u_t | \tilde{x}_t)$ that outputs the actual control action. Here $\phi \in \Phi, \psi \in \Psi$ represent the parameters for each network. In our implementation, both networks output the mean and diagonal covariance of multivariate Gaussian distributions. 

Computing the gradient of our objective \cref{eq:robust pomdp} is difficult due to the presence of the mutual information, so the MINE is used to approximate this term.\footnote{We note that, in order to employ the bound \cref{eq:cost bound}, we need to employ an overestimate of $\mathbb{I}[x_t; \tilde{x}_t]$. In \cref{sec:examples}, we demonstrate that in practice, minimizing the MINE is a practical approximation that produces robust policies.} This approximation allows us to derive a learning algorithm similar to policy gradient. Using the well-known identity
\begin{align}
	\nabla_\psi \pi^\psi(u_t | \tilde{x}_t) = \pi^\psi(u_t | \tilde{x}_t) \nabla_\psi \log \pi^\psi(u_t | \tilde{x}_t),
\end{align}
the gradient of the expected cost with respect to $\psi$ is given by
\begin{align}
	\nabla_\psi \EE\left[c(\tau)\right] &= \EE\left[\left(\sum_{t = 0}^{T} \nabla_\psi \log \pi^\psi(u_t | \tilde{x}_t)\right) \cdot c(\tau)\right]. \label{eq:grad psi}
\end{align}
Repeating this process for $\phi$ yields an analogous formula with $q^\phi(\tilde{x}_t | \tilde{x}_{t - 1}, y_t)$ replacing $\pi^\psi(u_t | \tilde{x}_t)$:
\begin{align}
	\nabla_\phi \EE\left[c(\tau)\right] &= \EE\left[\left(\sum_{t = 0}^{T} \nabla_\phi \log q^\phi(\tilde{x}_t | \tilde{x}_{t - 1}, y_t)\right) \cdot c(\tau)\right]. \label{eq:grad phi}
\end{align}

Finally, it remains to calculate $\nabla_{\phi, \psi} \hat{\mathbb{I}}^N[x_t; \tilde{x}_t]$. Though it is more tractable to optimize the MINE instead of the true mutual information, computing the gradient of the MINE with respect to the policy parameters is not entirely straightforward. The complexity lies in the fact that the gradient of the MINE with respect to these parameters depends on knowing or estimating the marginal distributions $p_t(x_t)$ and $q_t(\tilde{x}_t)$, both of which depend on $\phi$ and $\psi$. To approximate the MINE gradient we fix $\theta$ to the converged MINE parameters from \cref{alg:mine}, which yields:
\begin{align}
	\hat{\mathbb{I}}^N[x_t; \tilde{x}_t] = \hat{\EE}^N[F^\theta] - \log \hat{\EE}^N_\perp[\exp(F^\theta)]. \label{eq:mine theta fixed}
\end{align}
Since the neural networks used to parameterize the policy produce Gaussian distributions, we can represent $\tilde{x}_t$ as
\begin{align}
	\tilde{x}_t = \mu^\phi(\tilde{x}_{t - 1}, y_t) + A^\phi(\tilde{x}_{t - 1}, y_t)\epsilon_t, & & \epsilon_t \sim \mathcal{N}(0, I). \label{eq:lg}
\end{align}
This is similar to the reparameterization trick for variational autoencoders \cite{Kingma13}. Here, $\mu^\phi, A^\phi$ are computed using the mean and covariance output from the network $q^\phi(\tilde{x}_t | \tilde{x}_{t - 1}, y_t)$. Since the dynamics and sensor model are unknown, the approximations that $x_t$ and $y_t$ are independent of $\phi$ and $x_t, y_t, \tilde{x}_t$ are independent of $\psi$ are made. With these fixed, the gradient of \cref{eq:mine theta fixed} with respect to $\phi$ can be computed by storing $\epsilon_0, \dots, \epsilon_{t - 1}$ as a part of each rollout and backpropogating using \cref{eq:lg}.

The task-driven policy gradient (TDPG) algorithm is outlined in \cref{alg:trv-policy}. Our learning process involves training $T - 1$ MINE networks, whose parameters are denoted $\theta_0, \dots, \theta_{T - 1}$. For clarity in future sections, we refer to an iteration of the outer optimization loop as a policy epoch and an iteration of the optimization loop in \cref{alg:mine} as a MINE epoch. During each policy epoch, we rollout $N$ trajectories using the current policy parameters $\phi, \psi$. We then update each set of MINE parameters using \cref{alg:mine}. Once the MINE networks converge, they are used to approximate $\mathcal{J}$ and optimize the policy. This is repeated until the empirical estimate of $\mathcal{J}$, which is given by $\hat{\mathcal{J}} \coloneqq \beta \hat{\EE}^N\left[c(\tau)\right] + \sum_{t = 0}^T \hat{\mathbb{I}}^N[x_t; \tilde{x}_t]$ converges.

\textbf{Implementation Details.} It remains to specify how to select $\beta$ in a principled manner. Returning to the perspective described in \cref{sec:recurrent}, we treat $\mathcal{J}$ as the Lagrangian for minimizing the information shared between $x_t$ and $\tilde{x}_t$ subject to an upper bound on the maximum expected cost the policy is allowed. Then, we sweep through a set of values for $\beta$ and select the policy from the epoch with the lowest MINE estimate that also satisfies the specified limit on the empirical expected cost. This strategy produces the policy estimated to have the least state information present in the TRVs while satisfying our performance constraint.

It is likely that each MINE network is initialized to a poor estimate of the mutual information. In order to improve the initial estimate of the mutual information and its gradient, additional MINE epochs are used during the first policy epoch. In the following section, we will specify the number of additional epochs used. Moreover, as discussed in \cite{Belghazi18}, using minibatches to estimate the MINE gradient in \cref{eq:grad mine} leads to a biased estimate of the gradient. Replacing the denominator in \cref{eq:grad mine} with the exponential moving average (EMA) of its value compensates for this bias.\footnote{The EMA is a filter defined on a sequence $f_0, \dots, f_t$ by the recursive relationship $\hat{f}_{t} = (1 - \alpha) a_t + \alpha \hat{f}_{t - 1}, \hat{f}_0 = f_0$.} This technique is used in some examples in the following section. We also limit the policy networks to output Gaussian distributions with diagonal covariances.

\begin{algorithm}[t]
\caption{Task-Driven Policy Gradient (TDPG)}
\label{alg:trv-policy}
\begin{algorithmic}[1]
	\Repeat
	\State Rollout batch of $N$ trajectories: $\{(x_t^n, \tilde{x}_t^n, u_t^n)_{t = 0}^T\}_{n=0}^N$.
	\For{$t = 0, \dots, T - 1$}
	\State $\theta_t \gets $\textsc{Train-Mine}$(\theta_t, \{x_t^n, \tilde{x}_t^n)\}_{n = 1}^N)$\
	\EndFor
	\State Update $\phi, \psi$ using rollout batch and \cref{eq:grad psi}, \cref{eq:grad phi}, \cref{eq:lg}.
	\Until{Convergence of $\hat{\mathcal{J}}$.}
\end{algorithmic}
\end{algorithm}

\section{Examples}
\label{sec:examples}
In this section, we apply the algorithm described in \cref{sec:algorithm} to three problems: (i) a continuous state and action version of the ``Lava Problem" from the POMDP literature, (ii) a vision-based ball-catching example, and (iii) a grasping problem with depth-image observations. For each of these problems, we present thorough simulation results demonstrating that the task-driven policy is robust to distributional shifts in the sensor model and testing environment. We compare our method against a policy with the same parameterization as ours, but trained using a standard policy gradient method in order to minimize the expected cost associated with the problem.

\subsection{Lava Problem}
\label{ssec:lava}

\begin{figure}[t]
\begin{center}
\includegraphics[width=0.9\columnwidth]{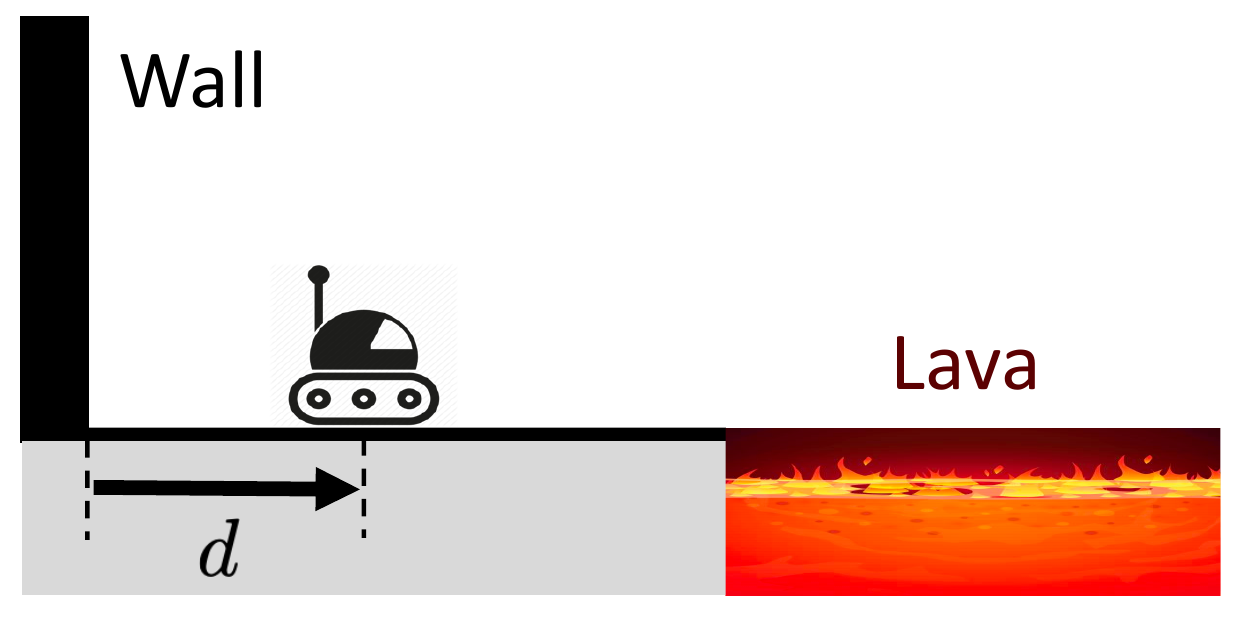}
\end{center} 
\vspace{-14pt}
\caption{An illustration of the Lava Problem described in \cref{ssec:lava}. The robot (a double integrator) needs to navigate to a target state without moving so far right that the robot falls into the lava. \label{fig:lava}}
\vspace{-18pt}
\end{figure}

The first problem we consider is a continuous state and action version of the Lava problem (Figure \ref{fig:lava}) \cite{Cassandra94, Florence2017, Kaelbling98, Pacelli19}, which is a common example for evaluating robust solutions to POMDPs. This scenario involves a robot navigating to a goal location along a line segment between a wall and a lava pit. The robot is modeled as a time-discretized double integrator, i.e. its state $x_t = \transp{[d_t\ v_t]}$ evolves with dynamics $x_{t + 1} = \transp{[d_t + v_t\ v_t + u_t]}$. Here $d_t$ is the displacement from the wall (in meters) and $v_t$ is the robot's velocity (in meters per second). The goal is to navigate to the state $g = \transp{[3\ 0]}$ within a time horizon of $T = 5$ steps. However, $d_t$ is limited to the interval $[0, 5]$m. If the robot collides with the wall located at $d = 0$, then its velocity is set to $0$m/s as well. If the robot's position exceeds $d_t = 5$m, then the robot falls into hot lava, where it is unable to move any further. At training time, the robot is provided with a high-quality estimate of its state. This is modeled by the choice $y_t \sim \mathcal{N}(x_t, \sigma^2 I)$, where $\sigma^2 = 0.0001$ and $\mathcal{N}$ denotes the Gaussian distribution. The cost function for this problem is $c_t(x_t, u_t) = \norm[2]{x_t - g}$ for $t = 0, \dots, T - 1$ and $c_T = 100\norm[2]{x_T - g}$. The robot is initialized with $v_t = 0$ and $d_t$ uniformly distributed between 0 and 5 meters.

\textbf{Training Summary.} Both $q^\phi(\tilde{x}_t | \tilde{x}_{t - 1}, y_t)$ and $\pi^\psi(u_t | \tilde{x}_t)$ have two hidden layers with 64 units and use exponential linear unit (ELU) nonlinearities \cite{Clevert15}. We choose a two-dimensional space $\tilde{\mathcal{X}}_t$ of TRVs. In this example, we found the performance of all learned policies improved when the parameters $\phi$ and $\psi$ were allowed to be time-varying. The MINE networks used in this example contain two hidden layers with 32 units each and ELU nonlinearities. A batch of 500 rollouts is used for each epoch, and a minibatch size of 50 is used for each MINE epoch, and an EMA with $\alpha = $5\textsc{e}-5 was used to compute the MINE gradient. The learning rates in this problem are 8\textsc{e}-4 for the policy networks and 5\textsc{e}-5 for the MINE network. All policies were trained for 300 epochs, with a computation time of about 10 seconds per epoch (about 50 minutes total for each policy). For this example, all computation was carried out on an Intel i9-7940X. Policies were trained with $\beta \in \{25^{-1}, 50^{-1}, 75^{-1}, 100^{-1}\}$ and the upper limit on the expected cost for selecting the policy was 40.

\begin{figure*}
\begin{center}
\subfigure[$\sigma^2 = 0.001$]{\includegraphics[trim=28 35 35 33, clip, width=0.245\linewidth]{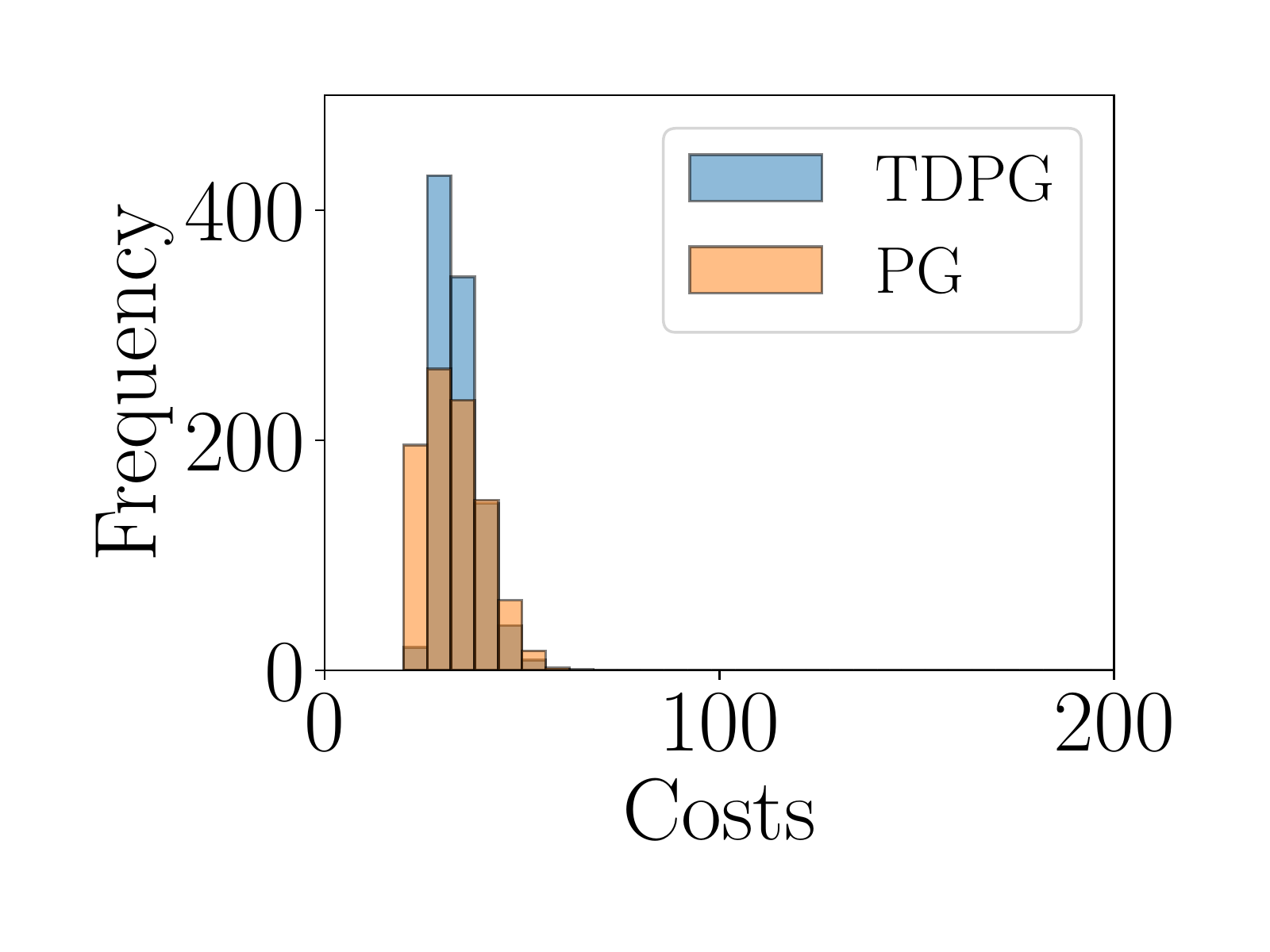}}
\subfigure[$\sigma^2 = 0.01$]{\includegraphics[trim=35 35 35 33, clip, width=0.245\linewidth]{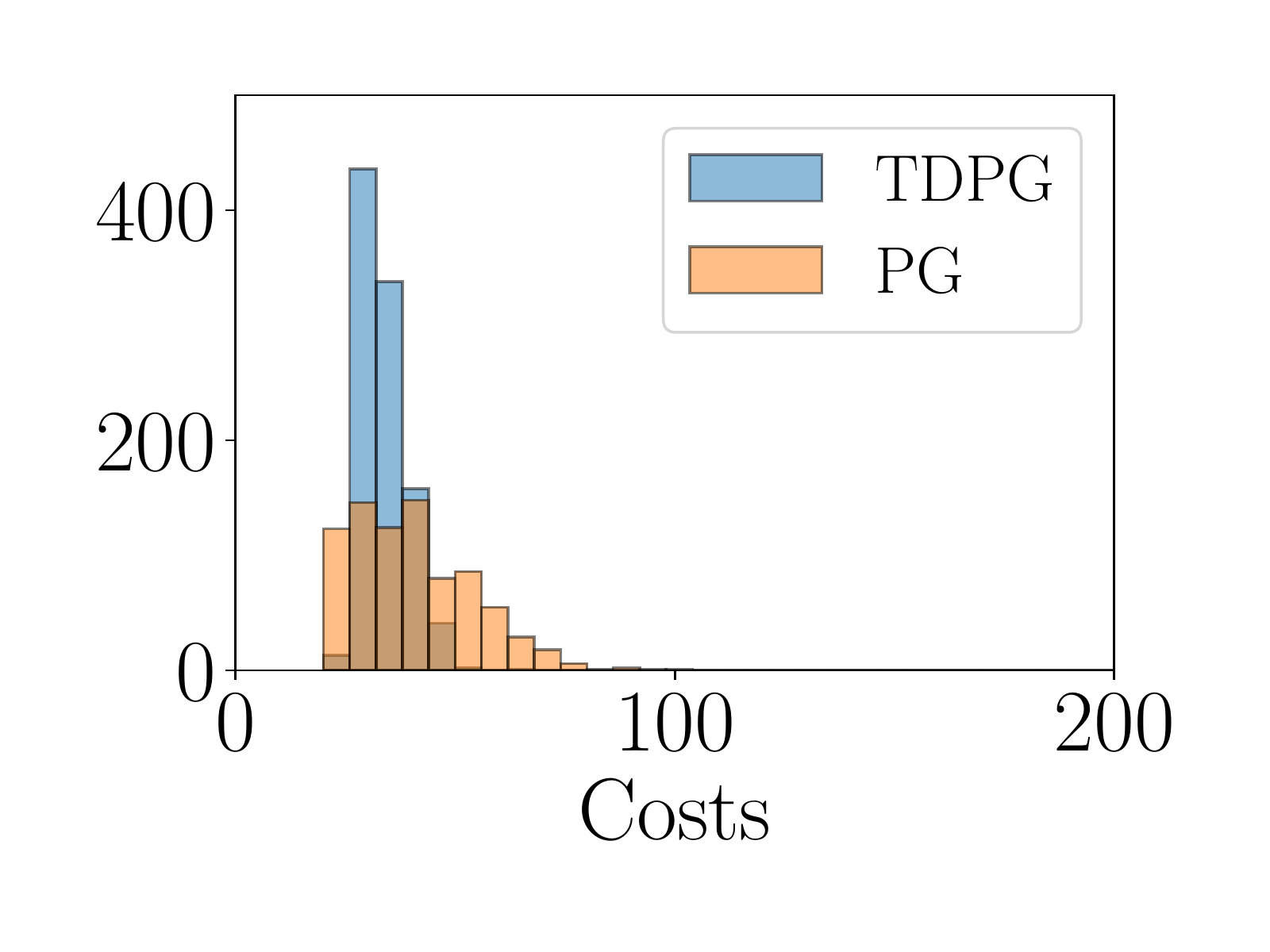}}
\subfigure[$\sigma^2 = 0.1$]{\includegraphics[trim=30 35 35 33, clip, width=0.245\linewidth]{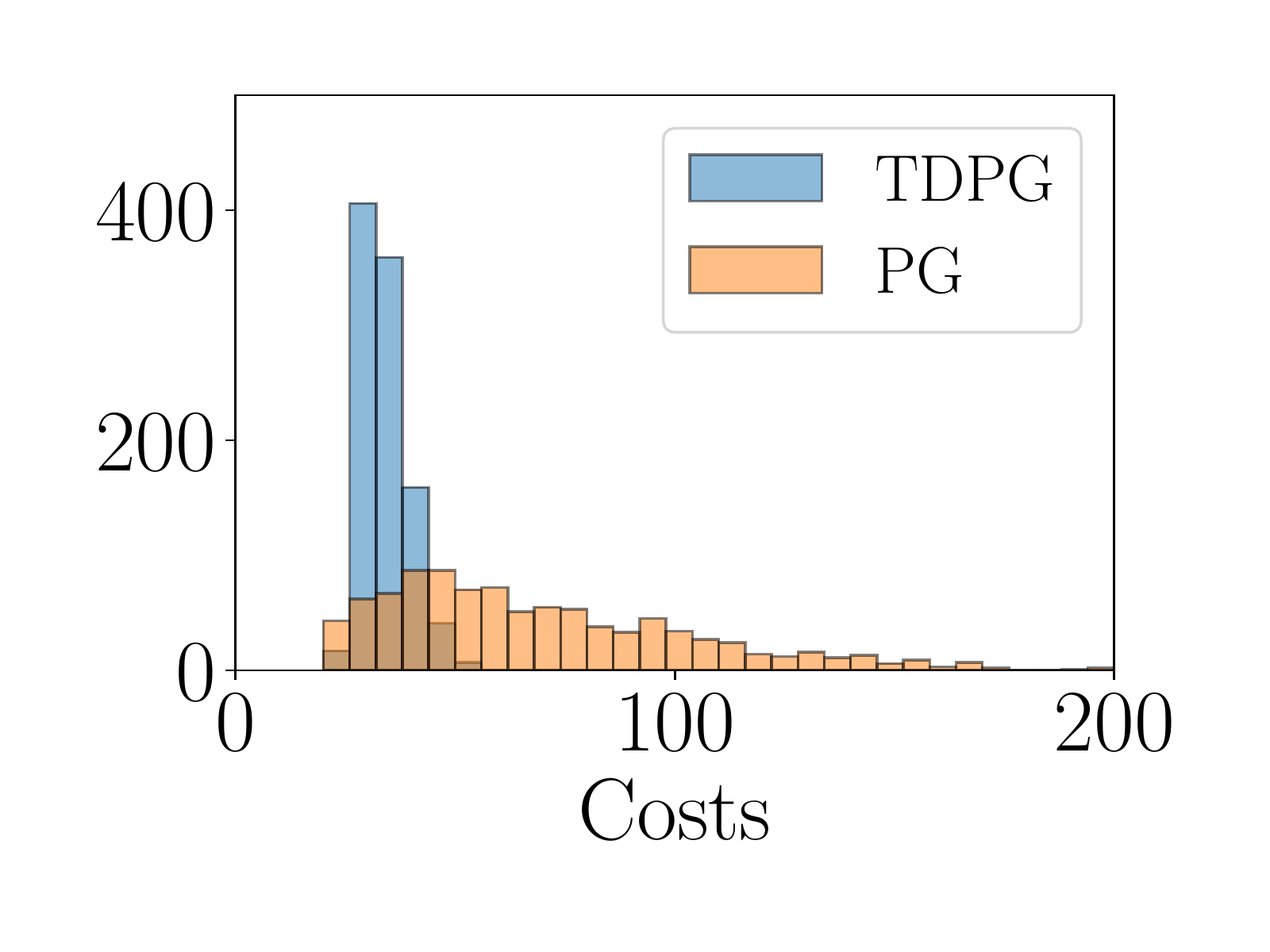}}
\subfigure[$\sigma^2 = 1$]{\includegraphics[trim=30 35 35 33, clip, width=0.245\linewidth]{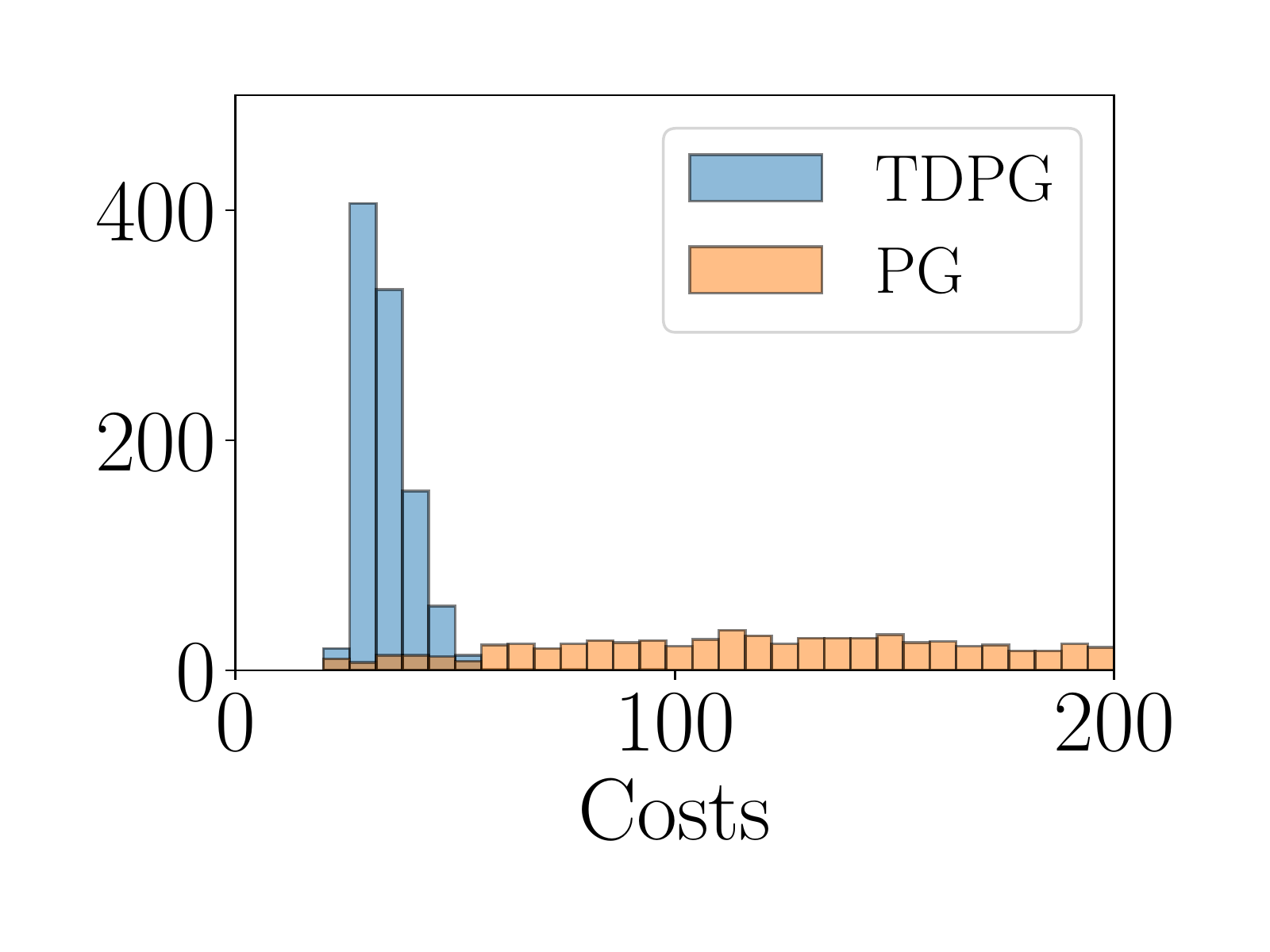}}
\vspace{-10pt}
\caption{These histograms compare our policy (TDPG) to the policy found by policy gradient (PG) in the Lava Problem  described in \cref{ssec:lava} by showing the frequencies at which each policy incurs different costs. The sensor noise of the test environment is increased from left to right. As the noise increases, the PG policy performance degrades, while the TDPG policy performance remains almost constant. This is because the TDPG algorithm found a task-relevant open-loop policy that is robust to this kind of disturbance. \label{fig:lava results}}
\end{center} 
\vspace{-18pt}
\end{figure*}

\begin{figure*}[!b]
\vspace{-15pt}
\centering
\subfigure[Training]{\includegraphics[width=0.115\linewidth]{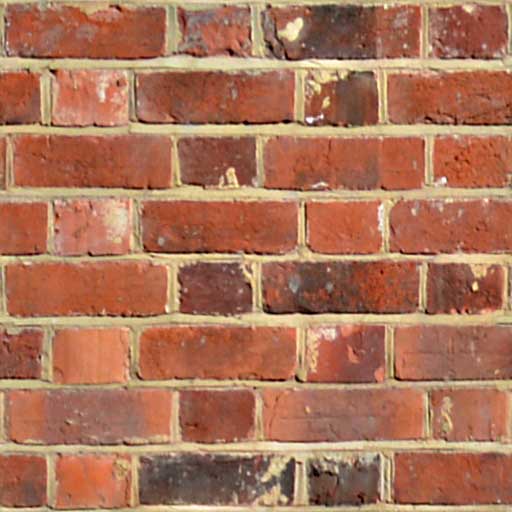}}
\subfigure[Test 1]{\includegraphics[width=0.115\linewidth]{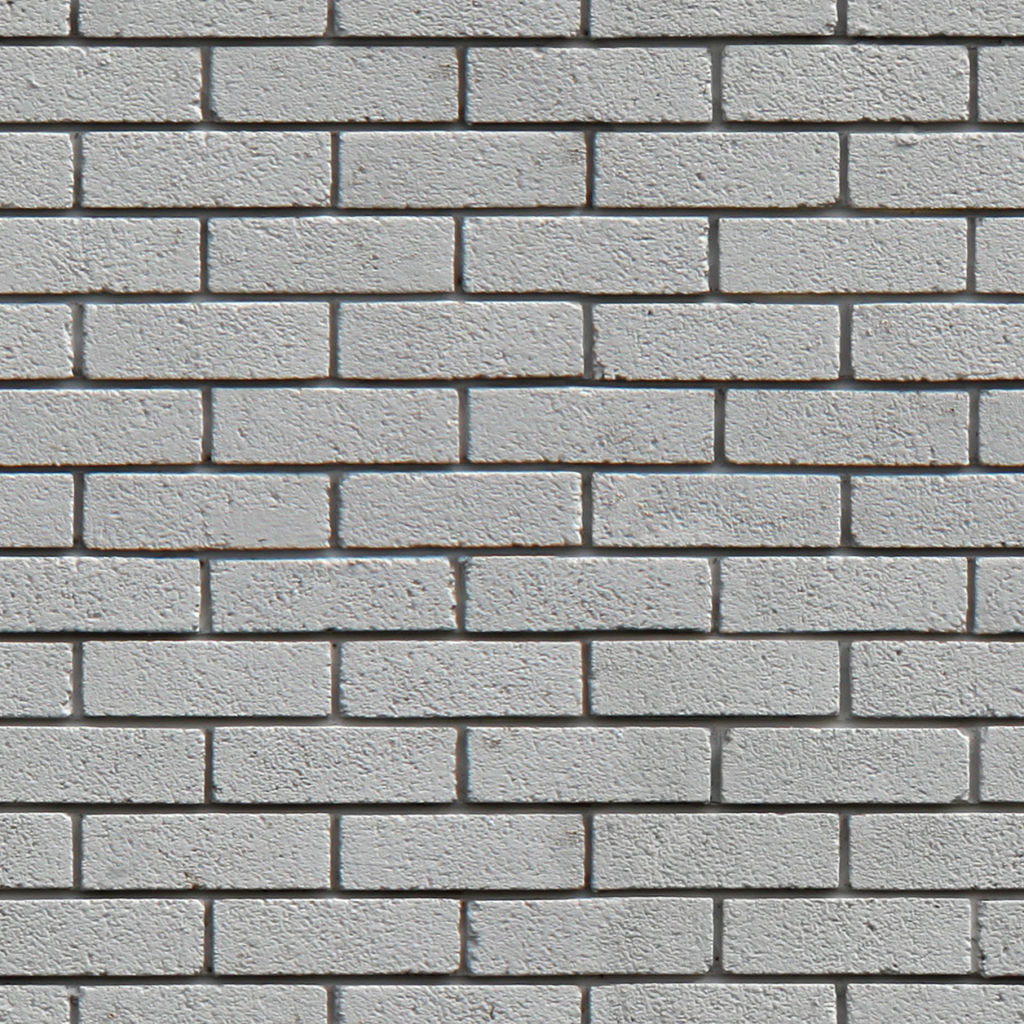}}
\subfigure[Test 2]{\includegraphics[width=0.115\linewidth]{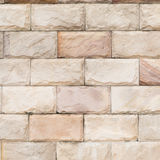}}
\subfigure[Test 3]{\includegraphics[width=0.115\linewidth]{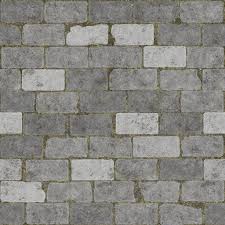}}
\subfigure[Test 4]{\includegraphics[width=0.115\linewidth]{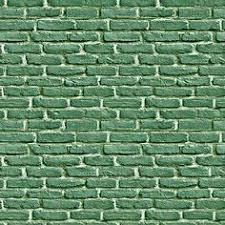}}
\subfigure[Test 5]{\includegraphics[width=0.115\linewidth]{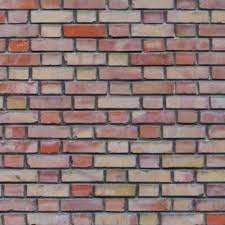}}
\subfigure[Test 6]{\includegraphics[width=0.115\linewidth]{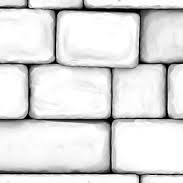}}
\subfigure[Test 7]{\includegraphics[width=0.115\linewidth]{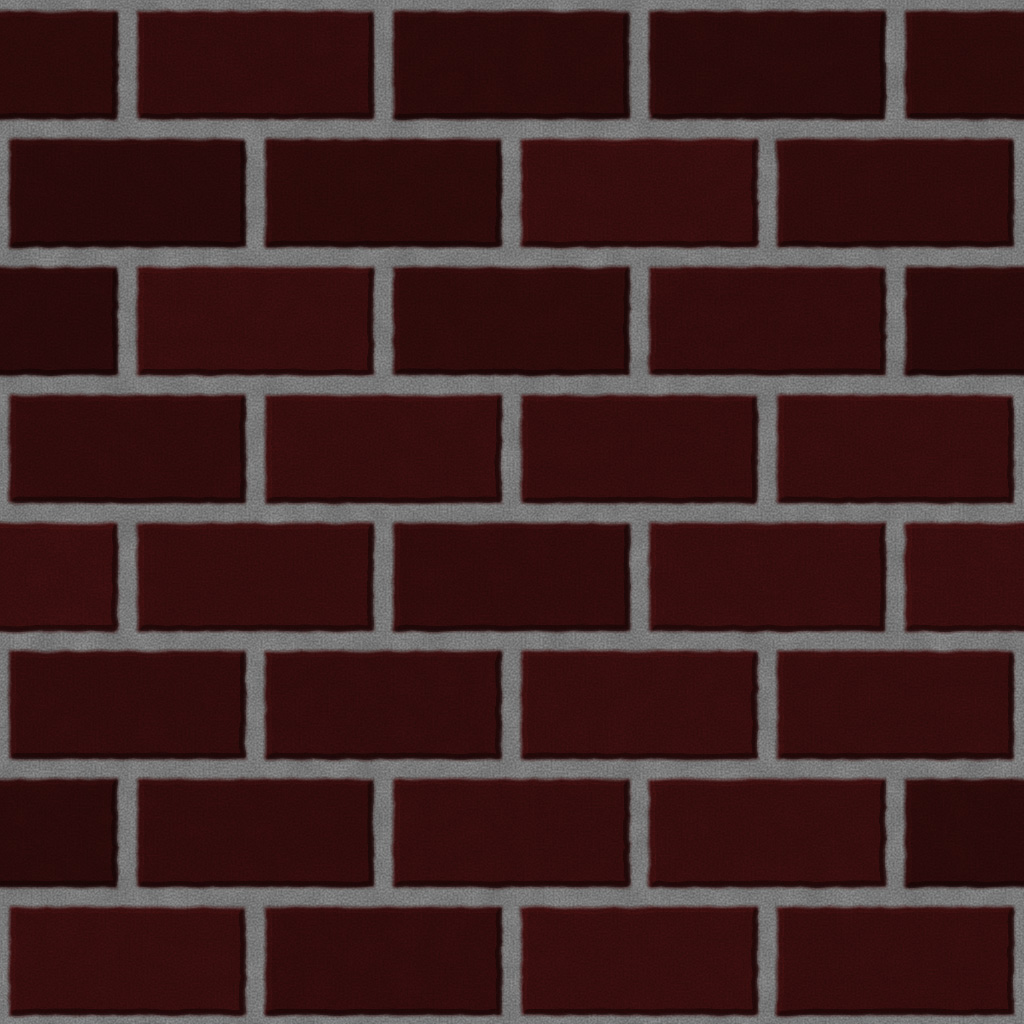}}
\caption{Textures used as backdrops in different ball-catching environments as part of the example described in \cref{ssec:ball}.}
\label{fig:textures}
\end{figure*}



\begin{table}[]
  \centering
  \begin{tabular}{|c|cc|cc|}
  	\hline
    Scenario & \multicolumn{2}{|c|}{Policy Gradient} & \multicolumn{2}{|c|}{Task-Driven PG}\\
    \hline
    & Mean & Std. & Mean & Std.\\
    \hline
    Training & \textbf{31.04} & \textbf{7.191} & 36.18 & 23.22 \\
    \hline
    \multicolumn{1}{|l|}{Sensor Noise $\sigma^2 = $1\textsc{e}-3} & \textbf{31.58} & \textbf{8.632} & 36.55 & 25.02 \\
    \multicolumn{1}{|l|}{Sensor Noise $\sigma^2 = $1\textsc{e}-2} & \textbf{35.69} & \textbf{15.68} & 36.00 & 22.32 \\
    \multicolumn{1}{|l|}{Sensor Noise $\sigma^2 = $1\textsc{e}-1} & 66.29 & 36.48 & \textbf{36.20} & \textbf{21.85} \\
    \multicolumn{1}{|l|}{Sensor Noise $\sigma^2 = $1\textsc{e}0} & 172.40 & 88.08 & \textbf{37.39} & \textbf{25.60} \\
    \hline
  \end{tabular}
  \caption{Performance of PG and TDPG Policies in Training and Testing Lava Environments}
  \label{tab:lava}
  \vspace{-10mm}
\end{table}

\textbf{Policy Evaluation.} We compare the resulting policy with one trained to minimize only the expected cost using policy gradient. The two policies are compared in \cref{fig:lava results}, and the statistics of their performance in different environments (corresponding to different levels of sensor noise) are presented in \cref{tab:lava} . Due to the presence of low sensor noise in the training environment, the PG policy finds it optimal to drive the robot directly towards the goal state. Interestingly, the TDPG algorithm finds a \emph{qualitatively different} strategy. In particular, TDPG recovers the robust open-loop behavior described in \cite{Cassandra94, Florence2017, Pacelli19}: regardless of initial position, the robot moves left until it collides with the wall, then moves right to the goal state.
As the sensor noise $\sigma^2$ is increased, the performance of the PG policy degrades rapidly. For example, if the sensor reports the robot is to the left of the goal when it is really to the right of the goal, it is likely to fall in the lava. In contrast, the TDPG is virtually unaffected by the increased sensor noise. 

\subsection{Vision-Based Ball Catching}
\label{ssec:ball}

Next, we consider a ball-catching example inspired by the gaze heuristic discussed in \cref{sec:intro} (see \cref{fig:anchor}). We formalize this problem by considering a ball confined to a plane with $x$ and $y$ coordinates $(b_t^x, b_t^y)$. The robot is confined to the $x$-axis and must navigate to $b_t^x$. The state of the system in this example is given by $x_t = \transp{[d_t\ b^x_t\ b^y_t\ v^x_t\ v^y_t]}$, where $d_t$ represents the robot's displacement along the $x$-axis, $v^x_t$ represents the ball's velocity along the $x$-axis, and $v^y_t$ represents the ball's velocity along the $y$-axis (i.e., the ball's vertical velocity). The dynamics are given by:
\begin{align}
	\begin{bmatrix}
	d_{t + 1}\\
	b_{t + 1}^x\\
	b_{t + 1}^y\\
	v_{t + 1}^x\\
	v_{t + 1}^y	
	\end{bmatrix} = 
	\begin{bmatrix}
	d_{t} + \Delta t \cdot u_t\\
	b_t^x + \Delta t \cdot v^x_t\\
	b_t^y + \Delta t \cdot v^y_t\\
	v_t^x\\
	v_t^y - \Delta t \cdot g	
	\end{bmatrix}.
\end{align}
Here, $g = 9.81 \mathrm{m}/\mathrm{s}^2$ is gravity, and $\Delta t = \frac{1}{15}\mathrm{s}$ is used to discretize the dynamics of the system in time. The robot's initial position is uniformly distributed over the interval $[-2, 2]$m. The ball is launched from $b^x_0 = 8$m and $b^y_0 = 1$m with fixed initial velocities $v_0^x = -4.5$m/s and $v_0^y = 7.85$m/s. These initial conditions are chosen such that the ball always spends a fixed number of time steps above the $x$-axis. The time horizon $T = 25$ is chosen such that $T$ is the last time step the ball remains above the $x$-axis. The cost function that we are trying to minimize is $c_t(x_t, u_t) = 0.01 \norm[2]{u_t}$ for $t = 0, \dots, T - 1$ and $c_T(x_T) = 100\norm[2]{d_T - b_T^x}$, which encourages the robot to be very close to the ball when it lands at the end of the time horizon. The sensor in this scenario is a camera mounted above the robot.
This camera provides $64 \times 64 $ RGB images with values scaled between 0 and 1. We also placed a wall with a red brick texture centered at 10m along the $x$-axis. All simulations are carried out using PyBullet \cite{Coumans18}. A sample observation from the camera is presented in \cref{fig:anchor}, and a video depicting both policies operating in training and testing environments for this scenario is available here: \href{https://youtu.be/Mwv0kkRveas}{https://youtu.be/Mwv0kkRveas}.

\textbf{Training Summary.} In this example, $q^\phi(\tilde{x}_t | \tilde{x}_{t - 1}, y_t)$ is parameterized by a network with 2 convolutional layers with 6 output channels, kernel size of 4, and stride of length 2, followed by two fully-connected layers using 32 units each. An ELU nonlinearity is applied between convolutional layers. After each fully-connected layer, $\tanh$ nonlinearities were used instead of ELU nonlinearities to prevent the values of $\tilde{x}_t$ from growing unbounded. The dimension of $\tilde{\mathcal{X}}_t$ is 8.  The network $\pi^\psi(u_t | \tilde{x}_t)$ contains a single linear layer. The MINE networks used in this example contain two hidden layers with 64 units each and ELU nonlinearities. A batch of 200 rollouts is used for each epoch, and a minibatch size of 20 is used to train the MINE network. The learning rates in this problem are 1\textsc{e}-3 for the policy networks and 5\textsc{e}-5 for the MINE network. When training the TDPG policy, it was initialized with the PG solution and trained for 100 policy epochs. Each policy epoch contained 100 MINE epochs with 100,000 MINE epochs used on the first policy epoch, and an EMA with $\alpha =$ 5\textsc{e}-5 was used to compute the MINE gradient. Policies with $\beta = 16^{-1}, 18^{-1}, \dots, 40^{-1}$ were evaluated, with the upper limit on the expected cost placed at 24. Rollouts were computed on an 3.7GHz i7-8700K CPU while all optimization was done using an Nvidia Titan Xp. Each policy epoch took about 45 seconds to compute.

\textbf{Policy Evaluation.} We consider two different kinds of testing environments. All statistics reported for test environments were computed using $1000$ rollouts. In the first group of examples, we add random noise to each pixel; the noise is sampled from a  Gaussian distribution $\mathcal{N}(0, \sigma^2)$, where $\sigma$ is varied between experiments. In order to ensure that the observed image is a valid RGB image after adding noise, we normalize the values to lie between 0 and 1. 
For each experiment, the mean cost and distance between the robot and the ball on the $x$-axis is presented in \cref{tab:ball catching}. As the level of image noise increases, the performance of the PG policy deteriorates dramatically while the TDPG policy's performance remains largely unchanged. 


In the second set of experiments, we qualitatively change the nature of the environment by changing the texture of the backdrop (i.e., the wall in the background). A representative portion of each texture is presented in \cref{fig:textures}. The TDPG policy outperforms the PG policy in each of these seven testing environments except background 7. This texture has a similar hue (red) to the texture in the training environment allowing PG to perform well. The TDPG policy, however, outperforms PG in environments whose textures consist of hues that are different than the ones seen during training. Together with the previous set of experiments, this result suggests that the TDPG policy is more robust to observations that are qualitatively different than those provided to the policy during training.


\begin{table}[]
  \centering
  \begin{tabular}{|c|cc|cc|}
  	\hline
    Scenario & \multicolumn{2}{|c|}{Policy Gradient} & \multicolumn{2}{|c|}{Task-Driven PG}\\
    \hline
    & Cost & Dist. (m) & Cost & Dist. (m) \\
    \hline
    Training & \textbf{8.702} & \textbf{0.085} & 18.577 & 0.184 \\
    \hline
    Sensor Noise $\sigma = 0.10$  & \textbf{19.12} & \textbf{0.189} & 19.48 & 0.193 \\
    Sensor Noise $\sigma = 0.15$ & 46.68 & 0.464 & \textbf{20.12} & \textbf{0.199} \\
    Sensor Noise $\sigma = 0.20$  & 124.63 & 1.244 & \textbf{21.59} & \textbf{0.214} \\
    Sensor Noise $\sigma = 0.25$  & 187.53 & 1.874 & \textbf{27.11} & \textbf{0.274} \\
    \hline
    Test Background 1 & 182.1 & 1.828  & \textbf{122.7} & \textbf{1.225} \\
    Test Background 2 & 214.3 & 2.141  & \textbf{79.90}  & \textbf{0.797} \\
    Test Background 3 & 138.7 & 1.385  & \textbf{26.62}  & \textbf{0.264} \\
    Test Background 4 & 95.38 & 0.952  & \textbf{28.11}  & \textbf{0.279} \\
    Test Background 5 & 82.01 & 0.818  & \textbf{36.95} & \textbf{0.367} \\
    Test Background 6 & 208.5 & 2.083  & \textbf{166.4}  & \textbf{0.166} \\
    Test Background 7 & \textbf{9.180} & \textbf{0.090}  & 14.92 & 0.147 \\
    \hline
  \end{tabular}
  \caption{Mean Cost and Final Distance of PG and TDPG Policies in Training and Testing Ball-Catching Environments}
  \label{tab:ball catching}
  \vspace{-15pt}
\end{table}

\subsection{Grasping using a Depth Camera}
\label{ssec:grasping}

\begin{table}[]
  \centering
  \begin{tabular}{|c|cc|cc|}
  	\hline
    Scenario & \multicolumn{2}{|c|}{Policy Gradient} & \multicolumn{2}{|c|}{Task-Driven PG}\\
    \hline
    & Mean & Std. & Mean & Std.\\
    \hline
    Training & 0.107 & 0.309 & 0.094 & 0.292 \\
    \hline
    Increased Rotation & 0.182 & 0.386 & \textbf{0.132} & \textbf{0.339} \\
    Increased Translation & 0.367 & 0.482 & \textbf{0.358} & \textbf{0.480} \\
    \hline
    \end{tabular}
  \caption{Performance of PG and TDPG Policies in Training and Testing Grasping Environments}
  \label{tab:grasping}
   \vspace{-10mm}
\end{table}

\begin{figure}
\begin{center}
\subfigure[Franka Emika Panda]{\includegraphics[trim=400 300 200 33, clip, width=0.49\linewidth]{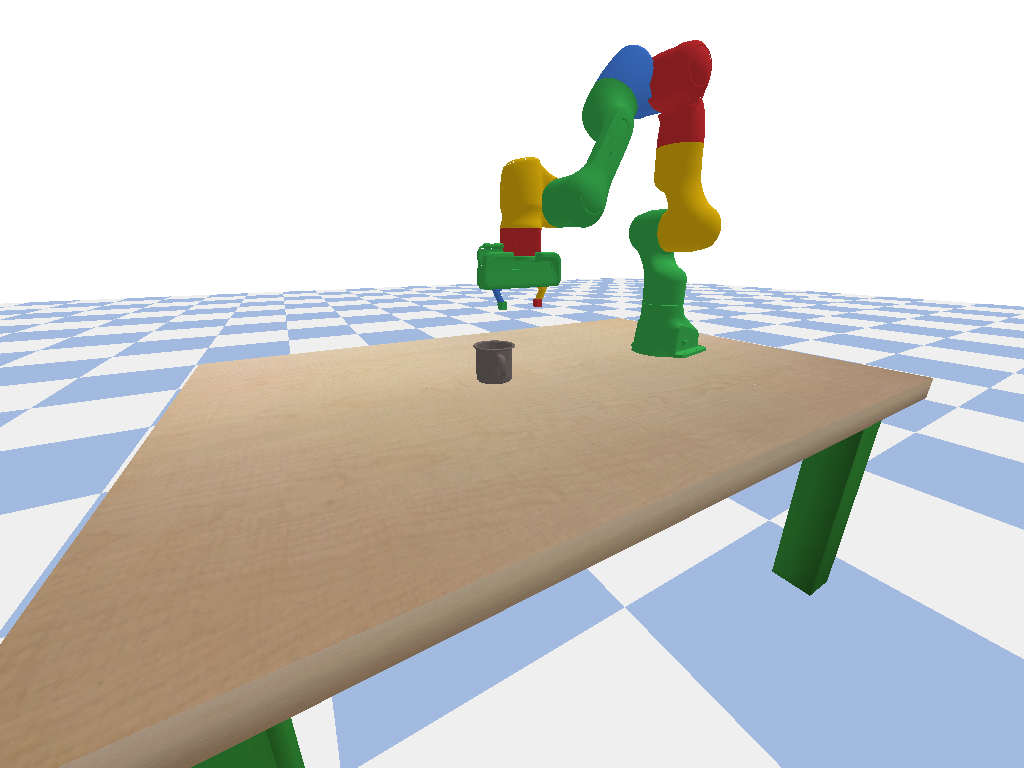}}
\subfigure[Sensor Observation]{\includegraphics[width=0.49\linewidth]{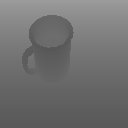}}

\caption{(a) A third-person perspective of the grasping scenario described in \cref{ssec:grasping} using PyBullet. (b) An upscaled example of the $128 \times 128$ depth image forming the sensor observation for this scenario. \label{fig:grasping}}
\end{center} 
\vspace{-28pt}
\end{figure}

In this final example, we consider the task of grasping and lifting an object (a mug) from a table using a Franka Emika Panda simulated with PyBullet (see \cref{fig:grasping}). This example is particularly interesting for studying task-driven policies due to the (approximate) radial symmetry of the mug about the vertical axis. Intuitively, this symmetry renders the orientation of the mug about the vertical axis largely irrelevant to the task of grasping the mug. This is due to the fact that the robot does not need to know the precise orientation of the mug; it only needs to know enough about the orientation in order to estimate if the handle will directly interfere with the gripper's location when the grasp is performed. 
We thus expect that a task-driven policy will remain largely unaffected by changes in the mug's orientation. Hence, a task-driven policy trained on a small set of mug rotations should generalize well to the full set of rotations. 

In this grasping problem, the state of the system $x_t \in \RR^4$ contains the position of the object and its rotation about the $z$-axis (which is oriented normal to the table surface) in radians. The control action $u_t \in \RR^4$ specifies the position and orientation of the end effector of the arm when the grasp occurs. After grasping, the arm attempts to move the end effector vertically. A cost of 0 is awarded if the arm successfully lifts the object more than 0.05 meters above the table and a cost of 1 is assigned otherwise. The observation is a 128 $\times$ 128 depth image. An example observation is included in \cref{fig:grasping}. The initial state of the object is sampled uniformly from the set $[0.45\ \mathrm{m}, 0.55\ \mathrm{m}] \times [-0.05\ \mathrm{m}, 0.05\ \mathrm{m}] \times \{0\ \mathrm{m}\} \times [-\frac{\pi}{4}, \frac{\pi}{4}]$.

\textbf{Training Summary.} We parameterized the policy in the following manner. The network $q^\phi(\tilde{x}_t | \tilde{x}_{t - 1}, y_t)$ contains two convolutional layers with 6 output channels. The first uses a kernel size of 6 and the second uses a kernel size of 4; both use a stride length of 2. The convolutional layers are followed by two fully-connected layers of sizes 128 and 64. An ELU linearity is used between convolutional layers and $\tanh$ nonlinearities are used after linear layers. The size of $\tilde{\mathcal{X}}_t$ is 16. The network $\pi(u_t | \tilde{x}_t)$ contains two fully connected layers with 64 units each and an ELU nonlinearity between them. The output of this network is then divided by 10 and added to a nominal control action of $[0.5, 0, 0.06, \frac{\pi}{6}]$. This scaling and translation is done to bias the policy to select grasping locations near the object to speed up learning early in the optimization process. Again, as in \cref{ssec:ball}, we initialize the TDPG solution to the PG solution at the start of training and learned policies with values of $\beta \in \{5^{-1}, 10^{-1}, 15^{-1}, 20^{-1}\}$ for 30 epochs.

In this example, the MINE proved particularly noisy and took longer to converge. To combat this, we increased the number of MINE epochs to 500,000 on the first policy epoch with 100,000 MINE epochs used on following policy epochs. To determine the value of the MINE, we applied an EMA with parameter $\alpha = 0.9$. No EMA was used for computing MINE gradients. The epoch with the lowest filtered MINE estimate with an expected cost below 0.15 was used for testing. Each epoch took approximately 5 minutes to compute. Rollouts were computed on an Intel 3.7GHz i7-8700K in parallel and optimization was done on a Titan Xp.

\textbf{Policy Evaluation.} The test results for this example are summarized in \cref{tab:grasping}. Again, all reported statistics are computed using 1000 trials. In the first testing environment, the set of angles the mug is placed at is expanded from $[-\frac{\pi}{4}, \frac{\pi}{4}]$ to $[0, 2\pi]$. The expected cost (i.e., grasp failure rate) of the PG policy increased twice as much as the TDPG policy in these new testing environments. In the second testing environment, the set of angles the object is placed at is again $[-\frac{\pi}{4}, \frac{\pi}{4}]$, but the $x$ and $y$ values were sampled from the larger set of $[0.425\ \mathrm{m}, 0.575\ \mathrm{m}] \times [-0.075\ \mathrm{m}, 0.075\ \mathrm{m}]$. In this setting, both policies perform equally poorly. This result supports our hypothesis that the rotation of the mug is largely unimportant for the task of lifting the mug because the TDPG policy was able to generalize to new mug angles, but not to the task-relevant translational coordinates of the mug. Meanwhile, the PG policy exhibits overfitting to task-irrelevant state information and is impacted poorly by its changes.

\section{Discussion and Conclusion} 
\label{sec:conclusion}
We presented a novel reinforcement learning algorithm for computing task-driven control policies for systems equipped with rich sensor observations (e.g., RGB or depth images). The key idea behind our approach is to learn a task-relevant representation that contains as little information as possible about the state of the system while being sufficient for achieving a low cost on the task at hand. Formally, this is achieved by using an information bottleneck criterion that minimizes the mutual information between the state of the system and a set of task-relevant variables (TRVs) used for computing control actions. We parameterize our policies using neural networks and present a novel policy gradient algorithm that leverages the recently-proposed mutual information neural estimator (MINE) for optimizing our objective. We refer to the resulting algorithm as task-driven policy gradient (TDPG).


We compare PG and TDPG policies in three experiments: an adaption of the canonical lava problem to continuous state spaces, a ball catching scenario inspired by the gaze heuristic from cognitive psychology, and a depth image-based grasping problem. In the lava example, the TDPG policy exploits the nonlinear dynamics to find a minimal information (open-loop) control policy that is robust to increased sensor noise. In the ball-catching example, the TDPG is also more robust to changes in the sensing model at test time than PG. These changes include both random noise corrupting the images and task-irrelevant structural changes, e.g. altering the textures in the robot's environment. Finally, in the grasping scenario, the TDPG policy generalizes to rotated states of the object not seen during training on which the PG policy struggles. Together, these scenarios validate that our approach to designing task-driven control policies produces robust policies that can operate in environments unseen during training.

\textbf{Future Work.} 
There are a number of challenges and exciting directions for further exploration. First, we have observed that MINE often results in noisy estimates of the mutual information and can take many epochs to converge. This results in long training times for TDPG. We also employed an approximation of the gradient of MINE with respect to policy parameters (due to the challenges associated with estimating state distributions at each time step, as described in Section \ref{sec:algorithm}). These observations motivate the exploration of other methods for minimizing the mutual information, eg., Stein variational gradient methods \cite{Haarnoja17, Liu16}. Another direction for future work is to adapt more advanced on-policy methods (e.g., proximal policy optimization (PPO) \cite{Schulman17}) to work with our approach, and potentially explore off-policy methods. We expect that these methods will be even more effective at learning task-driven policies. Finally, a particularly exciting direction for future work is to explore the benefits that our approach affords in terms of sim-to-real transfer. The simulation results in this paper suggest that TDPG can be more robust to sensor noise and perturbations to task-irrelevant features. Exploring whether this translates to more robust sim-to-real transfer is a promising direction we hope to explore in the future.



\begin{small}
\section*{Acknowledgements}

This work is partially supported by the National Science Foundation
[IIS-1755038], the Office of Naval Research [Award Number: N00014-18-1-2873], the Google Faculty Research Award, and the Amazon Research Award. We also gratefully acknowledge the support of NVIDIA Corporation with the donation of the Titan Xp GPU used for this research.
\end{small}

\bibliographystyle{abbrvnat}
\bibliography{irom}

\begin{thebibliography}{49}
\providecommand{\natexlab}[1]{#1}
\providecommand{\url}[1]{\texttt{#1}}
\expandafter\ifx\csname urlstyle\endcsname\relax
  \providecommand{\doi}[1]{doi: #1}\else
  \providecommand{\doi}{doi: \begingroup \urlstyle{rm}\Url}\fi

\bibitem[Achille and Soatto(2018)]{Achille18}
A.~Achille and S.~Soatto.
\newblock A separation principle for control in the age of deep learning.
\newblock \emph{Annual Review of Control, Robotics, and Autonomous Systems},
  1:\penalty0 287--307, 2018.

\bibitem[Alemi et~al.(2016)Alemi, Fischer, Dillon, and Murphy]{Alemi16}
A.~A. Alemi, I.~Fischer, J.~V. Dillon, and K.~Murphy.
\newblock Deep variational information bottleneck.
\newblock \emph{arXiv preprint arXiv:1612.00410}, 2016.

\bibitem[Anderson and Moore(2007)]{Anderson07}
B.~Anderson and J.~Moore.
\newblock \emph{Optimal Control: Linear Quadratic Methods}.
\newblock Courier Corporation, 2007.

\bibitem[Belghazi et~al.(2018)Belghazi, Baratin, Rajeshwar, Ozair, Bengio,
  Courville, and Hjelm]{Belghazi18}
M.~I. Belghazi, A.~Baratin, S.~Rajeshwar, S.~Ozair, Y.~Bengio, A.~Courville,
  and D.~Hjelm.
\newblock Mutual information neural estimation.
\newblock In \emph{Proceedings of the International Conference on Machine
  Learning}, pages 531--540, 2018.

\bibitem[Cassandra et~al.(1994)Cassandra, Kaelbling, and Littman]{Cassandra94}
A.~R. Cassandra, L.~P. Kaelbling, and M.~L. Littman.
\newblock Acting optimally in partially observable stochastic domains.
\newblock In \emph{AAAI}, volume~94, pages 1023--1028, 1994.

\bibitem[Clevert et~al.(2015)Clevert, Unterthiner, and Hochreiter]{Clevert15}
D.-A. Clevert, T.~Unterthiner, and S.~Hochreiter.
\newblock Fast and accurate deep network learning by exponential linear units
  ({ELU}s).
\newblock \emph{arXiv preprint arXiv:1511.07289}, 2015.

\bibitem[Coumans and Bai(2018)]{Coumans18}
E.~Coumans and Y.~Bai.
\newblock Pybullet, a python module for physics simulation for games, robotics
  and machine learning, 2018.

\bibitem[Cover and Thomas(2012)]{Cover12}
T.~M. Cover and J.~A. Thomas.
\newblock \emph{Elements of Information Theory}.
\newblock John Wiley \& Sons, 2012.

\bibitem[Dullerud and Paganini(2013)]{Dullerud13}
G.~E. Dullerud and F.~Paganini.
\newblock \emph{A Course in Robust Control Theory: A Convex Approach},
  volume~36.
\newblock Springer Science \& Business Media, 2013.

\bibitem[Florence(2017)]{Florence2017}
P.~R. Florence.
\newblock Integrated perception and control at high speed.
\newblock Master's thesis, Massachusetts Institute of Technology, 2017.

\bibitem[Gigerenzer(2007)]{Gigerenzer07}
G.~Gigerenzer.
\newblock \emph{Gut Feelings: The Intelligence of the Unconscious}.
\newblock Penguin, 2007.

\bibitem[Goyal et~al.(2019)Goyal, Islam, Strouse, Ahmed, Larochelle, Botvinick,
  Levine, and Bengio]{Goyal18}
A.~Goyal, R.~Islam, D.~Strouse, Z.~Ahmed, H.~Larochelle, M.~Botvinick,
  S.~Levine, and Y.~Bengio.
\newblock Transfer and exploration via the information bottleneck.
\newblock In \emph{Proceedings of the International Conference on Learning
  Representations}, 2019.

\bibitem[Gray(2011)]{Gray11}
R.~M. Gray.
\newblock \emph{Entropy and Information Theory}.
\newblock Springer Science \& Business Media, 2nd edition, 2011.

\bibitem[Gupta et~al.(2017{\natexlab{a}})Gupta, Davidson, Levine, Sukthankar,
  and Malik]{Gupta17}
S.~Gupta, J.~Davidson, S.~Levine, R.~Sukthankar, and J.~Malik.
\newblock Cognitive mapping and planning for visual navigation.
\newblock In \emph{Proceedings of the IEEE Conference on Computer Vision and
  Pattern Recognition}, pages 2616--2625, 2017{\natexlab{a}}.

\bibitem[Gupta et~al.(2017{\natexlab{b}})Gupta, Fouhey, Levine, and
  Malik]{Gupta17a}
S.~Gupta, D.~Fouhey, S.~Levine, and J.~Malik.
\newblock Unifying map and landmark based representations for visual
  navigation.
\newblock \emph{arXiv preprint arXiv:1712.08125}, 2017{\natexlab{b}}.

\bibitem[Haarnoja et~al.(2016)Haarnoja, Ajay, Levine, and Abbeel]{Haarnoja16}
T.~Haarnoja, A.~Ajay, S.~Levine, and P.~Abbeel.
\newblock Backprop {KF}: Learning discriminative deterministic state
  estimators.
\newblock In \emph{Advances in Neural Information Processing Systems}, pages
  4376--4384, 2016.

\bibitem[Haarnoja et~al.(2017)Haarnoja, Tang, Abbeel, and Levine]{Haarnoja17}
T.~Haarnoja, H.~Tang, P.~Abbeel, and S.~Levine.
\newblock Reinforcement learning with deep energy-based policies.
\newblock In \emph{Proceedings of the 34th International Conference on Machine
  Learning}, pages 1352--1361, 2017.

\bibitem[Jonschkowski et~al.(2018)Jonschkowski, Rastogi, and
  Brock]{Jonschkowski18}
R.~Jonschkowski, D.~Rastogi, and O.~Brock.
\newblock Differentiable particle filters: end-to-end learning with algorithmic
  priors.
\newblock In \emph{Proceedings of Robotics: Science and Systems}, Pittsburgh,
  Pennsylvania, June 2018.

\bibitem[Kaelbling et~al.(1998)Kaelbling, Littman, and Cassandra]{Kaelbling98}
L.~P. Kaelbling, M.~L. Littman, and A.~R. Cassandra.
\newblock Planning and acting in partially observable stochastic domains.
\newblock \emph{Artificial Intelligence}, 101\penalty0 (1-2):\penalty0 99--134,
  1998.

\bibitem[Karkus et~al.(2018)Karkus, Hsu, and Lee]{Karkus18}
P.~Karkus, D.~Hsu, and W.~S. Lee.
\newblock Particle filter networks: end-to-end probabilistic localization from
  visual observations.
\newblock \emph{arXiv preprint arXiv:1805.08975}, 2018.

\bibitem[Karkus et~al.(2019)Karkus, Ma, Hsu, Kaelbling, Lee, and
  Lozano-P{\'e}rez]{Karkus19}
P.~Karkus, X.~Ma, D.~Hsu, L.~P. Kaelbling, W.~S. Lee, and T.~Lozano-P{\'e}rez.
\newblock Differentiable algorithm networks for composable robot learning.
\newblock \emph{arXiv preprint arXiv:1905.11602}, 2019.

\bibitem[Kingma and Ba(2014)]{Kingma14}
D.~P. Kingma and J.~Ba.
\newblock Adam: A method for stochastic optimization.
\newblock \emph{arXiv preprint arXiv:1412.6980}, 2014.

\bibitem[Kingma and Welling(2013)]{Kingma13}
D.~P. Kingma and M.~Welling.
\newblock Auto-encoding variational bayes.
\newblock \emph{arXiv preprint arXiv:1312.6114}, 2013.

\bibitem[Klyubin et~al.(2005)Klyubin, Polani, and Nehaniv]{Klyubin05}
A.~S. Klyubin, D.~Polani, and C.~L. Nehaniv.
\newblock Empowerment: A universal agent-centric measure of control.
\newblock In \emph{IEEE Congress on Evolutionary Computation}, volume~1, pages
  128--135. IEEE, 2005.

\bibitem[Kolchinsky et~al.(2019)Kolchinsky, Tracey, and Wolpert]{Kolchinsky19}
A.~Kolchinsky, B.~D. Tracey, and D.~H. Wolpert.
\newblock Nonlinear information bottleneck.
\newblock \emph{Entropy}, 21\penalty0 (12):\penalty0 1181, 2019.

\bibitem[Levine et~al.(2016)Levine, Finn, Darrell, and Abbeel]{Levine16}
S.~Levine, C.~Finn, T.~Darrell, and P.~Abbeel.
\newblock End-to-end training of deep visuomotor policies.
\newblock \emph{The Journal of Machine Learning Research}, 17\penalty0
  (1):\penalty0 1334--1373, 2016.

\bibitem[Levine et~al.(2018)Levine, Pastor, Krizhevsky, Ibarz, and
  Quillen]{Levine18}
S.~Levine, P.~Pastor, A.~Krizhevsky, J.~Ibarz, and D.~Quillen.
\newblock Learning hand-eye coordination for robotic grasping with deep
  learning and large-scale data collection.
\newblock \emph{The International Journal of Robotics Research (IJRR)},
  37\penalty0 (4-5):\penalty0 421--436, 2018.

\bibitem[Liu and Wang(2016)]{Liu16}
Q.~Liu and D.~Wang.
\newblock Stein variational gradient descent: A general purpose bayesian
  inference algorithm.
\newblock In \emph{Advances in neural information processing systems}, pages
  2378--2386, 2016.

\bibitem[McLeod et~al.(2003)McLeod, Reed, and Dienes]{McLeod03}
P.~McLeod, N.~Reed, and Z.~Dienes.
\newblock Psychophysics: How fielders arrive in time to catch the ball.
\newblock \emph{Nature}, 426\penalty0 (6964):\penalty0 244, 2003.

\bibitem[Pacelli and Majumdar(2019)]{Pacelli19}
V.~Pacelli and A.~Majumdar.
\newblock Task-driven estimation and control via information bottlenecks.
\newblock In \emph{Proceedings of the IEEE International Conference on Robotics
  and Automation (ICRA)}, 2019.

\bibitem[Poole et~al.(2019)Poole, Ozair, Oord, Alemi, and Tucker]{Poole19}
B.~Poole, S.~Ozair, A.~v.~d. Oord, A.~A. Alemi, and G.~Tucker.
\newblock On variational bounds of mutual information.
\newblock \emph{arXiv preprint arXiv:1905.06922}, 2019.

\bibitem[Salge et~al.(2013)Salge, Glackin, and Polani]{Salge13}
C.~Salge, C.~Glackin, and D.~Polani.
\newblock Empowerment and state-dependent noise-an intrinsic motivation for
  avoiding unpredictable agents.
\newblock In \emph{Proceedings of the Artificial Life Conference}, pages
  118--125. MIT Press, 2013.

\bibitem[Salge et~al.(2014)Salge, Glackin, and Polani]{Salge14}
C.~Salge, C.~Glackin, and D.~Polani.
\newblock Empowerment--an introduction.
\newblock In \emph{Guided Self-Organization: Inception}, pages 67--114.
  Springer, 2014.

\bibitem[Schulman et~al.(2015)Schulman, Levine, Abbeel, Jordan, and
  Moritz]{Schulman15}
J.~Schulman, S.~Levine, P.~Abbeel, M.~Jordan, and P.~Moritz.
\newblock Trust region policy optimization.
\newblock In \emph{Proceedings of the International Conference on Machine
  Learning}, pages 1889--1897, 2015.

\bibitem[Schulman et~al.(2017)Schulman, Wolski, Dhariwal, Radford, and
  Klimov]{Schulman17}
J.~Schulman, F.~Wolski, P.~Dhariwal, A.~Radford, and O.~Klimov.
\newblock Proximal policy optimization algorithms.
\newblock \emph{arXiv preprint arXiv:1707.06347}, 2017.

\bibitem[Shaffer et~al.(2004)Shaffer, Krauchunas, Eddy, and McBeath]{Shaffer04}
D.~M. Shaffer, S.~M. Krauchunas, M.~Eddy, and M.~K. McBeath.
\newblock How dogs navigate to catch frisbees.
\newblock \emph{Psychological Science}, 15\penalty0 (7):\penalty0 437--441,
  2004.

\bibitem[Shapiro et~al.(2009)Shapiro, Dentcheva, and
  Ruszczy{\'n}ski]{Shapiro09}
A.~Shapiro, D.~Dentcheva, and A.~Ruszczy{\'n}ski.
\newblock \emph{Lectures on Stochastic Programming: Modeling and Theory}.
\newblock SIAM, 2009.

\bibitem[Soatto(2011)]{Soatto11}
S.~Soatto.
\newblock Steps towards a theory of visual information: Active perception,
  signal-to-symbol conversion and the interplay between sensing and control.
\newblock \emph{arXiv preprint arXiv:1110.2053}, 2011.

\bibitem[Soatto(2013)]{Soatto13}
S.~Soatto.
\newblock Actionable information in vision.
\newblock In \emph{Machine Learning for Computer Vision}, pages 17--48.
  Springer, 2013.

\bibitem[S{\"u}nderhauf et~al.(2018)S{\"u}nderhauf, Brock, Scheirer, Hadsell,
  Fox, Leitner, Upcroft, Abbeel, Burgard, Milford, and Corke]{Sunderhauf18}
N.~S{\"u}nderhauf, O.~Brock, W.~Scheirer, R.~Hadsell, D.~Fox, J.~Leitner,
  B.~Upcroft, P.~Abbeel, W.~Burgard, M.~Milford, and P.~Corke.
\newblock The limits and potentials of deep learning for robotics.
\newblock \emph{The International Journal of Robotics Research (IJRR)},
  37\penalty0 (4-5):\penalty0 405--420, 2018.

\bibitem[Sutton and Barto(1998)]{Sutton98}
R.~S. Sutton and A.~G. Barto.
\newblock \emph{Reinforcement Learning: An Introduction}, volume~1.
\newblock MIT press Cambridge, 1998.

\bibitem[Tiomkin et~al.(2017)Tiomkin, Polani, and Tishby]{Tiomkin17}
S.~Tiomkin, D.~Polani, and N.~Tishby.
\newblock Control capacity of partially observable dynamic systems in
  continuous time.
\newblock \emph{arXiv preprint arXiv:1701.04984}, 2017.

\bibitem[Tishby and Zaslavsky(2015)]{Tishby15}
N.~Tishby and N.~Zaslavsky.
\newblock Deep learning and the information bottleneck principle.
\newblock In \emph{Proceedings of the IEEE Information Theory Workshop}, pages
  1--5. IEEE, 2015.

\bibitem[Tishby et~al.(1999)Tishby, Pereira, and Bialek]{Tishby99}
N.~Tishby, F.~C. Pereira, and W.~Bialek.
\newblock The information bottleneck method.
\newblock In \emph{Proceedings of the 37th Allerton Conference on
  Communication, Control, and Computing}, 1999.

\bibitem[Yingjun and Xinwen(2019)]{Yingjun19}
P.~Yingjun and H.~Xinwen.
\newblock Learning representations in reinforcement learning: An information
  bottleneck approach.
\newblock \emph{arXiv preprint arXiv:1911.05695}, 2019.

\bibitem[Yu et~al.(2019)Yu, Shevchuk, Sadigh, and Finn]{Yu19}
T.~Yu, G.~Shevchuk, D.~Sadigh, and C.~Finn.
\newblock Unsupervised visuomotor control through distributional planning
  networks.
\newblock \emph{arXiv preprint arXiv:1902.05542}, 2019.

\bibitem[Zhao et~al.(2019)Zhao, Tiomkin, and Abbeel]{Zhao19}
R.~Zhao, S.~Tiomkin, and P.~Abbeel.
\newblock Learning efficient representation for intrinsic motivation.
\newblock \emph{arXiv preprint arXiv:1912.02624}, 2019.

\bibitem[Zhou and Doyle(1998)]{Zhou98}
K.~Zhou and J.~C. Doyle.
\newblock \emph{Essentials of Robust Control}, volume 104.
\newblock Prentice hall Upper Saddle River, NJ, 1998.

\bibitem[Zhu et~al.(2017)Zhu, Mottaghi, Kolve, Lim, Gupta, Fei-Fei, and
  Farhadi]{Zhu17}
Y.~Zhu, R.~Mottaghi, E.~Kolve, J.~J. Lim, A.~Gupta, L.~Fei-Fei, and A.~Farhadi.
\newblock Target-driven visual navigation in indoor scenes using deep
  reinforcement learning.
\newblock In \emph{Proceedings of the IEEE International Conference on Robotics
  and Automation (ICRA)}, pages 3357--3364. IEEE, 2017.

\end{thebibliography}

\end{document}